\documentclass[a4paper,conference]{IEEEtran}
\usepackage{lipsum}
\usepackage{times}
\usepackage{soul}
\usepackage{url}
\usepackage[hidelinks, backref]{hyperref}
\usepackage[utf8]{inputenc}
\usepackage[small]{caption}
\usepackage{graphicx}
\usepackage{cite}
\usepackage{comment}
\usepackage{amsthm}
\usepackage{booktabs}
\usepackage{algorithmicx,algorithm}
\usepackage[noend]{algpseudocode}
\usepackage{amsmath,amssymb,amsfonts}
\usepackage{graphicx}
\usepackage{textcomp}
\usepackage{xcolor}
\usepackage{url}            % simple URL typesetting
\usepackage{amsfonts}       % blackboard math symbols
\usepackage{nicefrac}       % compact symbols for 1/2, etc.
\usepackage{microtype}      % microtypography
\usepackage{mathrsfs}
\usepackage{amsmath}
\usepackage{amssymb}
\usepackage{graphicx, subfig}
\usepackage{epstopdf}
\usepackage{tikz}
\usepackage{pgf}
\usepackage{multirow}
\usepackage{pgfplots}
\usepackage{float}
\usepackage{pifont}
\newcommand\blfootnote[1]{%
  \begingroup
  \renewcommand\thefootnote{}\footnote{#1}%
  \addtocounter{footnote}{-1}%
  \endgroup
}
\ifCLASSINFOpdf
\else
\fi
\DeclareRobustCommand*{\IEEEauthorrefmark}[1]{%
    \raisebox{0pt}[0pt][0pt]{\textsuperscript{\footnotesize\ensuremath{#1}}}}

\begin{document}
%
% paper title
% Titles are generally capitalized except for words such as a, an, and, as,
% at, but, by, for, in, nor, of, on, or, the, to and up, which are usually
% not capitalized unless they are the first or last word of the title.
% Linebreaks \\ can be used within to get better formatting as desired.
% Do not put math or special symbols in the title.
\title{Progressive Cluster Purification \\ for Unsupervised Feature Learning}

% author names and affiliations
% use a multiple column layout for up to three different
% affiliations

\author{\IEEEauthorblockN{Yifei Zhang\IEEEauthorrefmark{1}\IEEEauthorrefmark{2}$^\ast$,
Chang Liu\IEEEauthorrefmark{3}$^\ast$,
Yu Zhou\IEEEauthorrefmark{1}$^\dagger$,
Wei Wang\IEEEauthorrefmark{1}\IEEEauthorrefmark{2}, Weiping Wang\IEEEauthorrefmark{1}, and
Qixiang Ye\IEEEauthorrefmark{3}$^\dagger$}
\IEEEauthorblockA{\IEEEauthorrefmark{1}Institute of Information Engineering, Chinese Academy of Sciences, Beijing, China}
\IEEEauthorblockA{\IEEEauthorrefmark{2}School of Cyber Security, University of Chinese Academy of Sciences, Beijing, China}
\IEEEauthorblockA{\IEEEauthorrefmark{3}University of Chinese Academy of Sciences, Beijing, China \\ \{zhangyifei0115, zhouyu, wangweiping\}@iie.ac.cn \\
\{liuchang615, wangwei1608\}@mails.ucas.ac.cn, qxye@ucas.ac.cn}
}

% conference papers do not typically use \thanks and this command
% is locked out in conference mode. If really needed, such as for
% the acknowledgment of grants, issue a \IEEEoverridecommandlockouts
% after \documentclass

% for over three affiliations, or if they all won't fit within the width
% of the page, use this alternative format:
%
%\author{\IEEEauthorblockN{Michael Shell\IEEEauthorrefmark{1},
%Homer Simpson\IEEEauthorrefmark{2},
%James Kirk\IEEEauthorrefmark{3},
%Montgomery Scott\IEEEauthorrefmark{3} and
%Eldon Tyrell\IEEEauthorrefmark{4}}
%\IEEEauthorblockA{\IEEEauthorrefmark{1}School of Electrical and Computer Engineering\\
%Georgia Institute of Technology,
%Atlanta, Georgia 30332--0250\\ Email: see http://www.michaelshell.org/contact.html}
%\IEEEauthorblockA{\IEEEauthorrefmark{2}Twentieth Century Fox, Springfield, USA\\
%Email: homer@thesimpsons.com}
%\IEEEauthorblockA{\IEEEauthorrefmark{3}Starfleet Academy, San Francisco, California 96678-2391\\
%Telephone: (800) 555--1212, Fax: (888) 555--1212}
%\IEEEauthorblockA{\IEEEauthorrefmark{4}Tyrell Inc., 123 Replicant Street, Los Angeles, California 90210--4321}}

% use for special paper notices
%\IEEEspecialpapernotice{(Invited Paper)}

% make the title area
\maketitle

% As a general rule, do not put math, special symbols or citations
% in the abstract
\begin{abstract}
In unsupervised feature learning, sample specificity based methods ignore the inter-class information, which deteriorates the discriminative capability of representation models. Clustering based methods are error-prone to explore the complete class boundary information due to the inevitable class inconsistent samples in each cluster. In this work, we propose a novel clustering based method, which, by iteratively excluding class inconsistent samples during progressive cluster formation, alleviates the impact of noise samples in a simple-yet-effective manner. Our approach, referred to as Progressive Cluster Purification (PCP), implements progressive clustering by gradually reducing the number of clusters during training, while the sizes of clusters continuously expand consistently with the growth of model representation capability. 
With a well-designed cluster purification mechanism, it further purifies clusters by filtering noise samples which facilitate the subsequent feature learning by utilizing the refined clusters as pseudo-labels. Experiments on commonly used benchmarks demonstrate that the proposed PCP improves baseline method with significant margins. 
$\blfootnote{$^*$ Equal contribution\\ $^{\dagger}$ Corresponding authors\\}$
Our code will be available at \url{https://github.com/zhangyifei0115/PCP}.

\end{abstract}

% no keywords

% For peer review papers, you can put extra information on the cover
% page as needed:
% \ifCLASSOPTIONpeerreview
% \begin{center} \bfseries EDICS Category: 3-BBND \end{center}
% \fi
%
% For peerreview papers, this IEEEtran command inserts a page break and
% creates the second title. It will be ignored for other modes.
\IEEEpeerreviewmaketitle

\section{Introduction}
Representation learning has achieved unprecedented success in various computer vision tasks. This can be attributed to the availability of large-scale datasets with precise label annotations and convolutional neural networks (CNNs) capable of absorbing the annotation information~\cite{MinEntropy2019,FreeAnchor2019}. Nevertheless, precisely annotating samples on large-scale datasets is laborious, expensive, or even impractical.

As an alternative method, unsupervised feature learning (UFL), which is usually based on data completeness, sample distribution and instance similarity, has attracted increased attention. The completeness-based methods typically leverage the intrinsic structure of the input data by forcing models to predict the hidden part in specific well-designed pretext tasks, such as context predicting \cite{Context_Doersch_2015_ICCV}, colorization \cite{Color_Guatav_2016_ECCV}, jigsaw puzzles \cite{Jigsaw_Noroozi_2016_ECCV},
counting \cite{counting_2017_ICCV}, split-brain \cite{splitbrain_2017_CVPR} and rotations prediction \cite{Rotate_Gidaris_2018_ICLR}. 
The distribution-based methods could use Auto-Encoders \cite{autoencoder}, Variational Auto-Encoders \cite{kingma2013autoencoding} and Generative Adversarial Nets \cite{GAN-NIPS2014} to approximate the distribution of raw data. 
Both methods focus on intra-class distributions but unfortunately ignore inter-class information of samples that the discriminative capacity of extracted features is not fully explored. 

\begin{figure*}
\centering
% width=1.0\columnwidth
\includegraphics[width=1.6\columnwidth]{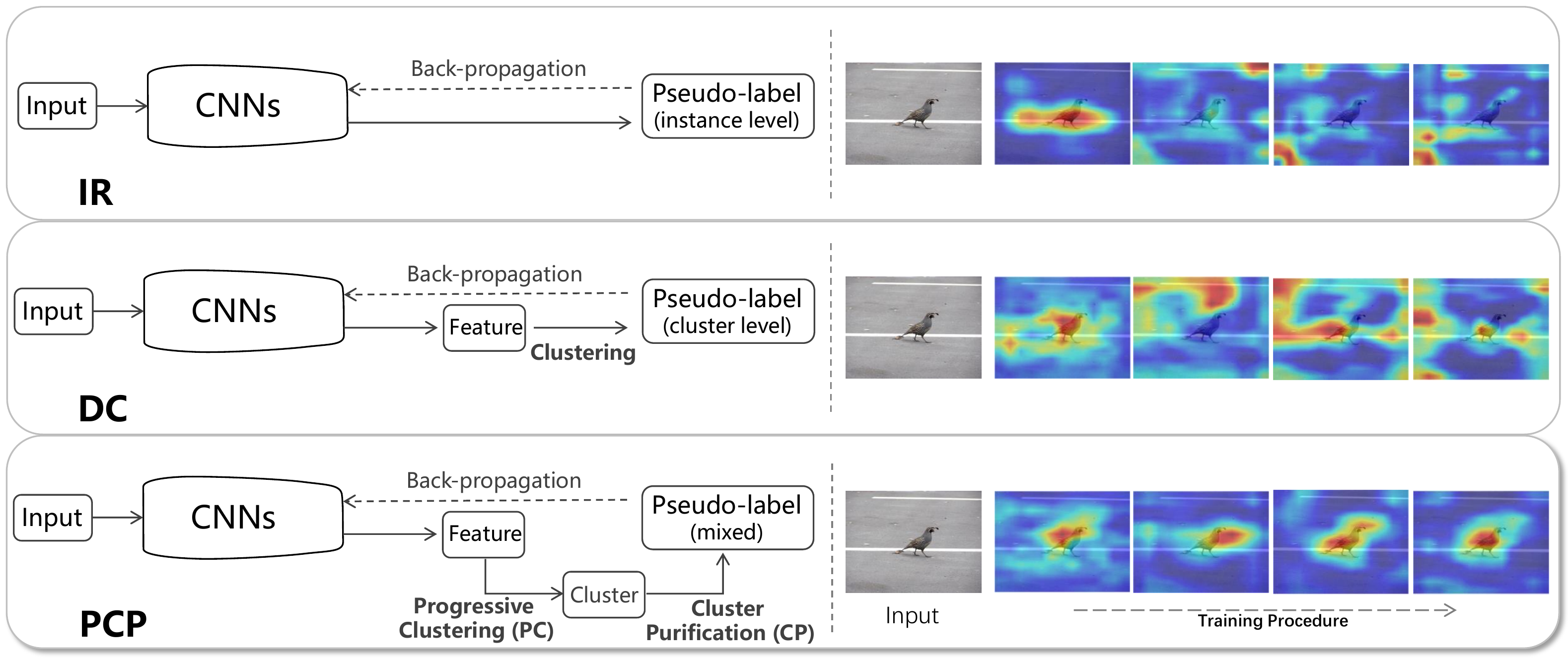}
\caption[]{ Comparison of unsupervised feature learning frameworks including IR~\cite{Wu_2018_CVPR}, DC~\cite{Caron_2018_ECCV} and our PCP (left) and visualization of corresponding features extracted by them (right). The features learned by IR and DC incorporate more background noise while features by our PCP are more stable and consistent with the foreground information.
}
\label{fig:motivation}
\end{figure*}

As one of the most representative similarity-based UFL method, DeepCluster (DC)~\cite{Caron_2018_ECCV} iteratively assigns samples to clusters according to their similarity in the feature space and then uses the assignment as pseudo category labels to train a feature extractor.
However, DC is puzzled by a significant number of class inconsistent instances (noise samples) within clusters, which could seriously deteriorate the representation learning performance. 
To avoid the impact of clustering noise, instance recognition (IR) \cite{Wu_2018_CVPR} treats each sample as an independent class and achieves impressive performance on commonly used benchmarks. 
Nevertheless, it totally misses the intrinsic inter-sample class information, which limits its discriminative capability of learned features. 
To improve discriminative capability, the anchor neighbourhood discovery (AND) \cite{Huang_2019_ICML} method adopts a divide-and-conquer strategy to find local neighbours under the constraint of cluster consistency. 
However, the sample noise problem remains not systematically solved, which makes the learned features less representative. 

In this study, we propose a simple-yet-effective UFL method, termed Progressive Cluster Purification (PCP), to alleviate the impact of sample noise and estimate reliable sample labels. PCP roots in simple $k$-means clustering while introducing a Progressive Clustering (PC) strategy and a Cluster Purification (CP) module to refine the clustering assignments. With iterative training, reliable class consistent samples converge together so that clear and purified pseudo labels are well estimated while discriminative features are learned in a progressive self-guided manner. As shown in Fig.~\ref{fig:motivation}, IR, DC and our PCP take three different strategy to learn discrimimative features in unsupervised manner, where PCP shows its superiority in more stable and efficient training procedure and richer feature representation capability.

Progressive Clustering is implemented by reducing the number of clusters from the number of samples towards the number of categories during the multiple iterations of clustering, so that the size of each cluster progressively increases. In this way, intra-cluster variance increases consistently with the growth of model representation capability so that the class inconsistent samples could be controlled at a reasonable level, which drives the model to improve the discriminability of extracted features. 

Progressive Clustering reduces the noise pollutants from the sources. Meanwhile, Cluster Purification is applied on clustering results in each epoch, with the aim to explicitly obtain stable and reliable clusters. CP consists of two noise processing procedure including unreliable sample filtering (CP$_r$) and unstable sample filtering (CP$_s$). With unreliable sample filtering, instances far away from cluster centroids are identified as noise samples, each of which is temporarily regarded as a distinct class. Thereafter, in order to improve the robustness of CP, we further introduce unstable sample filtering, a voting strategy, which is utilized to reassign class consistent instances based on the prevenient filtering results. PCP avoids network focus on low-level feature for clustering by warm-up training strategy. With PCP, the obtained pseudo labels contain purified and clear class discriminative information that facilitates the successive feature learning procedure.  

The contributions of this work are summarized as follows: 
\begin{enumerate}
		\item We propose the simple-yet-effective Progressive Cluster Purification (PCP) method, which implements progressive clustering (PC) by iteratively expanding cluster sizes and progressively enforcing feature discrimination power.

		\item A novel cluster purification (CP) mechanism is designed to exclude class inconsistent instances, which further aggregates the discrimination power of features and speeds up the convergence of unsupervised learning.
		
		\item With warm-up training strategy and well designed objective function, PCP improves the baseline method with significant margins on commonly used benchmarks.
\end{enumerate}

%---------------------------------------------------------------------------------
\begin{figure*}[!t]
     \centering
     \includegraphics[width=1.6\columnwidth]{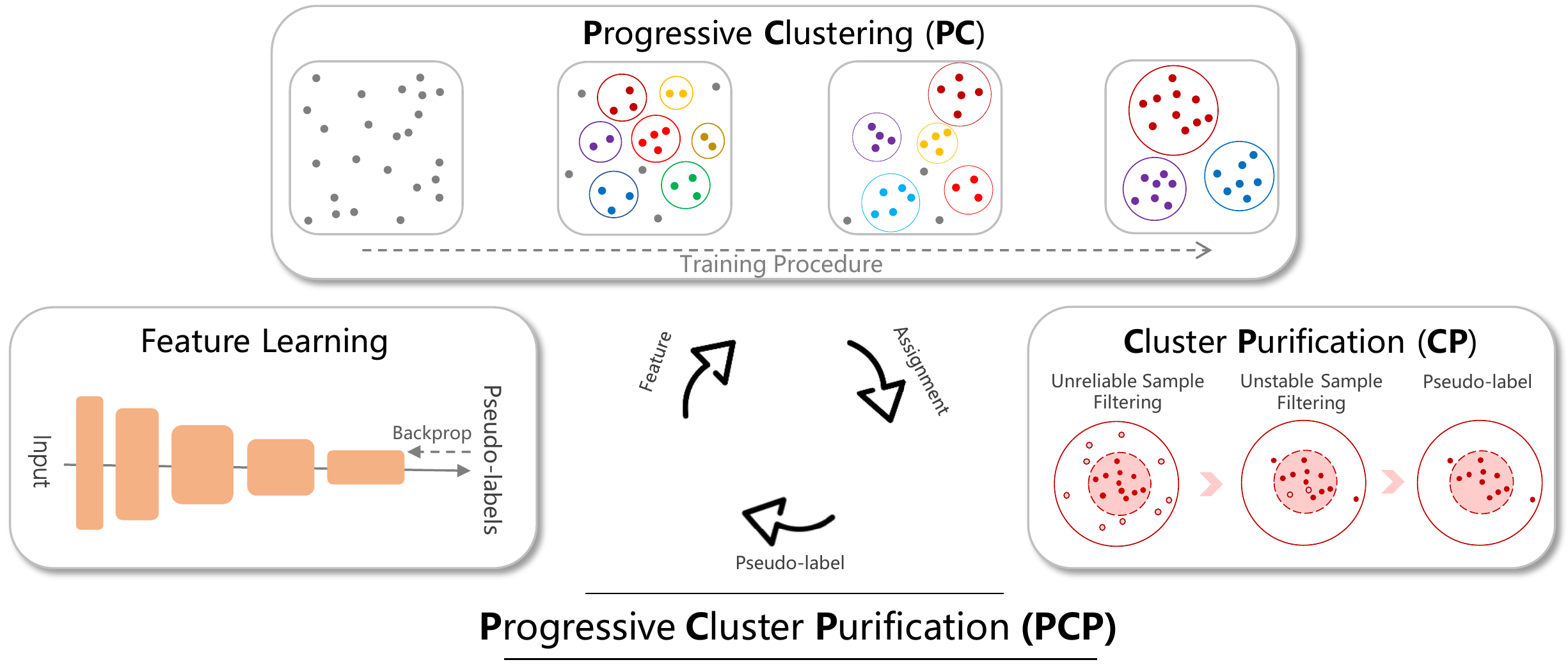}
     \caption{Overview of our PCP for unsupervised feature learning. 
     {Progressive Clustering (PC)} is implemented by gradually reducing the number of clusters while continuously expanding the size of clusters during training.
     {Cluster Purification (CP)} targets at improving reliability and stability of clustering by identifying unreliable instances and unstable instances. {Feature Learning} is carried out by using assignments as supervision. }
     \label{fig:pipline}
 \end{figure*}

\section{Related Works}

From a general perspective, unsupervised feature learning methods can be classified into three categories: completeness-based methods, distribution-based methods and similarity-based methods.

{\bf Completeness-based methods.} 
This kind of methods usually follows self-supervision, which designs pretext tasks by hiding some information of examples to construct a prediction task. By training models to predict the hidden information, rich feature representation could be learned.
Pretext tasks include context predicting \cite{Context_Doersch_2015_ICCV}, colorization \cite{Color_Guatav_2016_ECCV}, jigsaw puzzles \cite{Jigsaw_Noroozi_2016_ECCV}, counting \cite{counting_2017_ICCV}, split-brain \cite{splitbrain_2017_CVPR} and rotations \cite{Rotate_Gidaris_2018_ICLR}.
After that, FeatureDecoupling~\cite{Feng_2019_CVPR} proposes combining rotation and its unrelated part, VCP \cite{luo2020video} proposes to fill the blank with video options and PRP~\cite{PRP-CVPR20} produces self-supervision
signals about video playback rates. 
Although completeness-based methods can obtain proper feature representation for some specific tasks, it has a low performance bound as the learning procedure focuses on optimizing pretext tasks. In addition, it is unclear whether these pretext tasks are suitable for downstream tasks or not in new domains.

{\bf Distribution-based methods.}
The main idea of distribution-based methods is using encoders, $e.g.$, restricted Boltzmann machines  \cite{Hinton_2006_NC} or auto-encoders \cite{autoencoder}, to reconstruct data distribution while learning feature representation. 
Deep feature encoders such as deep belief networks  \cite{Hinton_2006_NC}, deep Boltzmann machines \cite{dbm_2009}, variational auto-encoders~\cite{kingma2013autoencoding} and generative adversarial network~\cite{GAN-NIPS2014} have been explored for feature learning. Despite the effectiveness, these methods unfortunately ignore inter-class information of samples thus the discriminative power of features is not fully explored. 

{\bf Similarity-based methods.}
In the deep learning era, similarity-based feature learning can be implemented by associating clustering with CNNs~\cite{Coates2012,Yang_2016_CVPR,Caron_2018_ECCV,LA_Zhuang_2019_ICCV}. As a representative method, DC~\cite{Caron_2018_ECCV} iteratively groups features and uses the assignments as pseudo labels to train feature representation. 
DC leverages the advantages of self-supervised learning and clustering to make up their disadvantages. 
%LA \cite{LA_Zhuang_2019_ICCV} propose the method of local aggregation in feature space. 
To alleviate the impact of noise samples, Exemplar CNN~\cite{Examplar2016} and IR \cite{Wu_2018_CVPR} treat each sample as an independent class. Recent invariant information clustering \cite{IIC_Ji_2019_ICCV} learns a neural network classifier from scratch with solely unlabelled data samples, and the objective is to maximize mutual information~\cite{CMC,mmi_2019_NIPS} between the class assignments of each pair. AND \cite{Huang_2019_ICML} adopts a divide-and-conquer method to find local neighbours under the constraint of cluster consistency.

Similarity-based methods, $e.g.$ DC, have set strong baselines for UFL. However, the significant performance gap between UFL and supervised feature learning indicates that it is a challenging task when representation models and sample clusters require to be estimated simultaneously. Existing methods remain challenged by the significant number of noise samples during clustering, which deteriorate the performance of representation models.

\section{Progressive Cluster Purification}

In unsupervised feature learning, clustering based method is susceptible to noisy supervision caused by inevitable class inconsistent samples. However, its superiority in reasoning class boundaries, which is so called class conceptualization, should not be neglected. 
Inspired by this, we propose the Progressive Cluster Purification approach targeting at alleviating the negative propagation of the noisy supervision during feature learning. 
As shown in Fig.~\ref{fig:pipline}, PCP consists of three components including progressive clustering, cluster purification and feature learning, which cooperate with each other in a circular way. 
PC assigns samples to distinct clusters in the feature space constructed during feature learning and the assignments are purified by CP to generate subsequent supervision for feature learning. 

Given an imagery data set $X = {\{x_{1}, x_{2}, x_{3}, \cdots, x_{n}\}}$ without any class label, pseudo-labels produced by PCP are used as supervision to derive CNN with parameters $\theta$. The learned model maps an input image $x_{i}$ to a feature space $\mathcal{V}(t)$ at the $t$-th epoch, as $v_{i}(t) = f_{\theta_{t}}(x_{i})$. Details about the main components of the proposed PCP are illustrated in the following paragraph.

\subsection{Progressive Clustering}
IR \cite{Wu_2018_CVPR} evidences that neural networks, to some extent, can learn class discriminative feature representation with solely instance-level supervision which regards individual instances as distinct classes. 
Inspired by this, we would go one more step to infer that deep models can extract the underlying class information under different grain-level supervision from instance-wise to class-wise. More importantly, in early phase of feature learning, the network has not converged that clustering with few centroids would cooperate significant class inconsistent samples within each cluster. With contaminated pseudo labels, inferior class conceptualization will further hinder the model representation capability growth.

To alleviate this situation, we propose progressive clustering to improve unsupervised class conceptualization by shrinking the number of clusters from the number of samples towards the number of true categories, Fig.~\ref{fig:pipline}. 
Consistently with the growth of our model’s representation capability, the clusters gradually expand while less noise sample pollutants occur.
As the total of samples $N$ is tremendous, we implement a linear declining strategy on its logarithm to decide the number of clusters at the $t$-th epoch denoted as $N_{t}$, which is formulated as

\begin{equation}
lg(N_{t}) = (1 - \frac{t}{T})lg(N) ,
\label{decline}
\end{equation}
where $T$ denotes the total of training epochs (200 as default). As shown in Eq.(\ref{decline}), $N_{t}$ declines fast in early epochs and goes slower in later epochs. When $t = 0$, $N_{t}$ equals to $N$. As the true category number is usually unknown prior knowledge, when $t \geq t_{0}$, we set $N_{t}$ fixed as $N_{t_{0}}$ that the cluster number $N_{t}$ begins with $N$ and stops at $N_{t_{0}}$ (hyperparameter). Table. \ref{tab:table3}
% and Fig.~\ref{fig:comparison} (b) 
show that our PCP is not as sensitive to the prior knowledge of the cluster number as DC~\cite{Caron_2018_ECCV} does.

\subsection{Cluster Purification}
With PC, the feature space becomes compact and more class-consistent, which facilitates class conceptualization. To further make the cluster more reliable and stable to optimize the feature learning under supervision of clear pseudo-labels with less noise, we design a Cluster Purification (CP) mechanism which consists of unreliable sample filtering (CP$_r$) and unstable sample filtering (CP$_s$). 

\textbf{Unreliable Sample Filtering (CP$_r$).} Based on the observation that samples near cluster centroids share higher apparent similarity, thus they are more likely to belong to the same class. We discard the samples far away from the centroids and temporarily regard each one as a distinct class in the subsequent learning procedure.
As shown in the lower right of Fig.~\ref{fig:pipline}, the pink area within the dashed circle represents the reliable district and hollow points indicate samples to be excluded while solid points to remain. We denote the unreliable sample filtering ratio in each cluster as $\gamma$ that with higher $\gamma$, few samples remain in a cluster. When $\gamma$ is close to 1, PCP degenerates to IR. That is to say, we build a bridge between sample specificity based methods and clustering based methods through $\gamma$ that essentially makes our PCP absorb the advantages of clear pseudo supervision with more inter-class information.
In the feature space $\mathcal{V}(t)$, we denote the set of reliable class consistent samples of $c$-th cluster, $\mathcal{S}_{c}(t)$, as $\mathcal{S}^{r}_{c}(t)$ and the corresponding set of noise samples to be discarded as $\mathcal{N}^{r}_{c}(t)$. It is obvious that $\mathcal{S}_{c}(t) = \mathcal{S}^{r}_{c}(t) \cup \mathcal{N}^{r}_{c}(t)$. Thereafter, these preliminary purification results of CP$_r$ are fed the successive purification procedure.

\textbf{Unstable Sample Filtering (CP$_s$).} Easily distinguished samples are likely to be consistently assigned to the same cluster at different iterations of clustering. However, perplexing samples go the opposite that clustering assignments of which are inconsistent and unstable.
Inspired by this, we propose a voting function $V(v_{i}(t), v_{j}(t))$ which utilizes the previous clustering results to quantitatively estimate the class consistency of samples $v_{i}(t)$ and $v_{j}(t)$ in the same cluster from the feature space $\mathcal{V}(t)$. For cluster $S_{c}(t)$, we denote the sample closest to the centroid as $v_{i_{c}}(t)$. Thereafter, the voting score of sample $v_{i}(t)$ from cluster $S_{c}(t)$, according to the progressive clustering result in the past $n$ epochs from the $t$-th epoch, can be calculated as

\begin{equation}
V(v_{i}(t), v_{i_{c}}(t)) = \sum_{k=0}^{n}\alpha^{k} \cdot {\bf \delta}(C(v_{i}(t-k)),  C(v_{i_{c}}(t-k))) ,
    \label{VNF}
\end{equation}
where $\alpha \in (0, 1)$ denotes the decay rate, $C(v_{i}(t))$ denotes the pseudo label which sample $v_{i}(t)$ belongs to, and ${\bf \delta}(x, y) = 1$ when $x = y$, otherwise -$1$. 

%If $V(i, i_{c})$ is lower than a threshold $\beta_1$ and $i \in I_{c}^{s}(t)$, sample $i$ is discarded as a noise sample, and it will belong to set $I_{c}^{n}(t)$, otherwise if $V(i_{c}, i)$ is higher than a threshold $\beta_2$ and $i \in I_{c}^{n}(t)$, sample $i$ is pulled back as a class-consistent sample, and it will belong to set $I_{c}^{s}(t)$, which are denoted as hollow points ($I_{c}^{n}(t)$) and solid points ($I_{c}^{s}(t)$) respectively shown in the lower right of Fig.~\ref{fig:pipline}.
As shown in the lower right of Fig.~\ref{fig:pipline}, we set different thresholds in $\mathcal{S}^{r}(t) = \cup_c \mathcal{S}_{c}^{r}(t)$ and $\mathcal{N}^{r}(t) = \cup_c \mathcal{N}_{c}^{r}(t)$ to discard samples in $\mathcal{S}^{r}(t)$ with low voting scores (hollow points) and pull back samples in $\mathcal{N}^{r}(t)$ with high voting scores (solid points). Then sets $\mathcal{S}_{c}^{r}(t)$ and $\mathcal{N}_{c}^{r}(t)$ are refined as $\mathcal{S}_{c}^{s}(t)$ and $\mathcal{N}_{c}^{s}(t)$ respectively. $\mathcal{S}^{s}(t) = \cup_c \mathcal{S}_{c}^{s}(t)$ and $\mathcal{N}^{s}(t) = \cup_c \mathcal{N}_{c}^{s}(t)$ are regarded as mixed instance-level and cluster-level pseudo labels for feature learning.

%----------------------------------------------
\begin{algorithm}[t] 
\caption{Progressive Cluster Purification.}
\label{algorithm1} 
\hspace*{0.02in} {\bf Input:}
{An imagery dataset $X$ without labels;} \\
\hspace*{0.02in} {\bf Output:}
{CNN model $f_{\theta}$ with parameters ${\theta}$ ;}
\begin{algorithmic}[1]
% \State PC: Progressive clustering, CP$_r$: Unreliable filtering, CP$_s$: Unstable filtering;
\State Preset embedding feature dimension $D$, training epochs $T$, cluster number $N_{t_0}$ for stopping declining;
\For{epoch $t$ = 1 {\bf to} $T$}
    % \State Get $N_t = max(N_{t},N_{t_0})$ during {\bf PC}, Eq.(\ref{decline});
    % \State Obtain feature space $\mathcal{V}(t)$ by $v(t) = f_{\theta_{t}}(x)$;
    % \State Implement clustering algorithm to get $\cup_{c} \mathcal{S}_{c}(t)$;
    % \For{each cluster $c$ = 1 {\bf to} $N_t$}
    %     \State Split $\mathcal{S}_{c}(t)$ into $\mathcal{S}_{c}^{r}(t)$ and $\mathcal{N}_{c}^{r}(t)$ by {\bf CP$_r$};
    %     \State Update $\mathcal{S}_{c}^{s}(t)$ and $\mathcal{N}_{c}^{s}(t)$ by {\bf CP$_s$}, Eq.(\ref{VNF});
    % \EndFor
    % \State  Calculate objective loss $L_{pcp}^{t}$ (Eq.(\ref{loss})) according to $\cup_{c} \mathcal{S}_{c}^{s}(t)$ and $\cup_{c} \mathcal{N}_{c}^{s}(t)$;
    % % \State Calculate embedding feature $V = f_{\theta}(X)$ and objective loss, Eq.(\ref{loss}) and Eq.(\ref{loss_warm});
    % \State Feature learning by updating model weights;
    \State Get the number of clusters $N_t = max(N_{t},N_{t_0})$ during the process of {\bf PC}, Eq.(\ref{decline});
    \State Obtain $D$-dimensional feature space $\mathcal{V}(t)$ by CNN model, $v(t) = f_{\theta_{t}}(x)$;
    \State Implement $k$-means clustering algorithm to get $\cup_{c} \mathcal{S}_{c}(t)$ with $N_t$ clusters;
    \For{each cluster $c$ = 1 {\bf to} $N_t$}
        \State Split $\mathcal{S}_{c}(t)$ into class consistent set $\mathcal{S}_{c}^{r}(t)$ and noise set $\mathcal{N}_{c}^{r}(t)$ by {\bf CP$_r$};
        \State Update class consistent set as $\mathcal{S}_{c}^{s}(t)$ and noise set as $\mathcal{N}_{c}^{s}(t)$ by {\bf CP$_s$}, Eq.(\ref{VNF});
    \EndFor
    \State  Calculate objective loss $L_{pcp}^{t}$ (Eq.(\ref{loss})) according to the union of set, $\cup_{c} \mathcal{S}_{c}^{s}(t)$ and $\cup_{c} \mathcal{N}_{c}^{s}(t)$;
    % \State Calculate embedding feature $V = f_{\theta}(X)$ and objective loss, Eq.(\ref{loss}) and Eq.(\ref{loss_warm});
    \State Feature learning by gradient back-propagation and updating model weights;
\EndFor
\State \Return $f_{\theta}$.
\end{algorithmic}
\end{algorithm}

\subsection{Feature Learning}
%An imagery data set $X = {\{x_{1}, x_{2}, x_{3}, \cdots, x_{n}\}}$ without any class label is used to derive CNN with parameters $\theta$ under the supervision of pseudo-labels processed by cluster purification (CP). The learned model maps an input image $x_{i}$ to a feature space $\mathcal{V}(t)$ at the $t$-th epoch, as $v_{i}(t) = f_{\theta_{t}}(x_{i})$.
IR \cite{Wu_2018_CVPR} involves a non-parametric softmax classifier aiming to minimize the negative log-likelihood objective function on the training set where each sample is regarded as a distinct class. The probability of an input $x$ being recognized
as $i$-th example is

\begin{equation}
    P(i|v) = \frac{exp(v_{i}^{T}v/\tau)}{\sum_{j=1}^{n}exp(v_{j}^{T}v/\tau)} ,
    \label{p-ir}
\end{equation}
where $v = f_{\theta}(x)$ and $\tau$ is a temperature parameter used for controlling the distribution concentration degree \cite{hinton_distilling}.
Following IR, we develop our instance-wise supervision loss and cluster-wise supervision loss at $t$-th epoch as

\begin{equation}
    L_{instance}^t = - \sum_{i \in \mathcal{N}^{s}(t)}log(P(i|v_{i}(t))) ,
    \label{linstance}
\end{equation}
and
\begin{equation}
    L_{cluster}^t = -{\sum_{c = 1}^{N_{t}} \sum_{i,j \in \mathcal{S}_{c}^{s}(t)}  log(P(i|v_{j}(t)))} .
    \label{lcluster}
\end{equation}
%where $I^{n}(t) = \cup_{c} (I_{c}^{n}(t))$ denotes the set of noise samples and $I_{c}^{s}(t)$ the set of class consistent samples of cluster $I_{c}(t)$.
With Eq.(\ref{linstance}) and Eq.(\ref{lcluster}), we can formally propose our loss function as

\begin{equation}
    L_{pcp}^{t} = L_{instance}^{t} + L_{cluster}^{t} .
    \label{loss}
\end{equation}

So far, the learning procedure of PCP can be summarized as Algorithm \ref{algorithm1}.

{\bf Warm-up.} 
In early training epochs, the cluster number dramatically decreases, which challenges the stability and convergence of feature learning. In addition, early clustering may lead to network focus on low-level features. To alleviate this situation, a warm-up training strategy is implemented by adding a branch sharing the same architecture and weights with the original PCP branch to assist its early learning. Following IS~\cite{Ye_2019_CVPR}, the branch is implemented in a sample specific learning manner aiming at learning random data augmentation invariant and instance spread-out features. With warm-up for training a certain epochs, the additional IS branch is discarded for further improvement of the PCP branch. Figures in Table~\ref{tab:table3} and Table~\ref{tab:table4+} clearly show the effect of the warm-up training strategy.

%----------------------------------------------

\subsection{Discussion}
To show the advantages of PCP over existing UFL methods, we compare them from three aspects including reliability, stability, and efficiency.

\begin{figure}[!t]
     \centering
     \includegraphics[width=1.0\columnwidth]{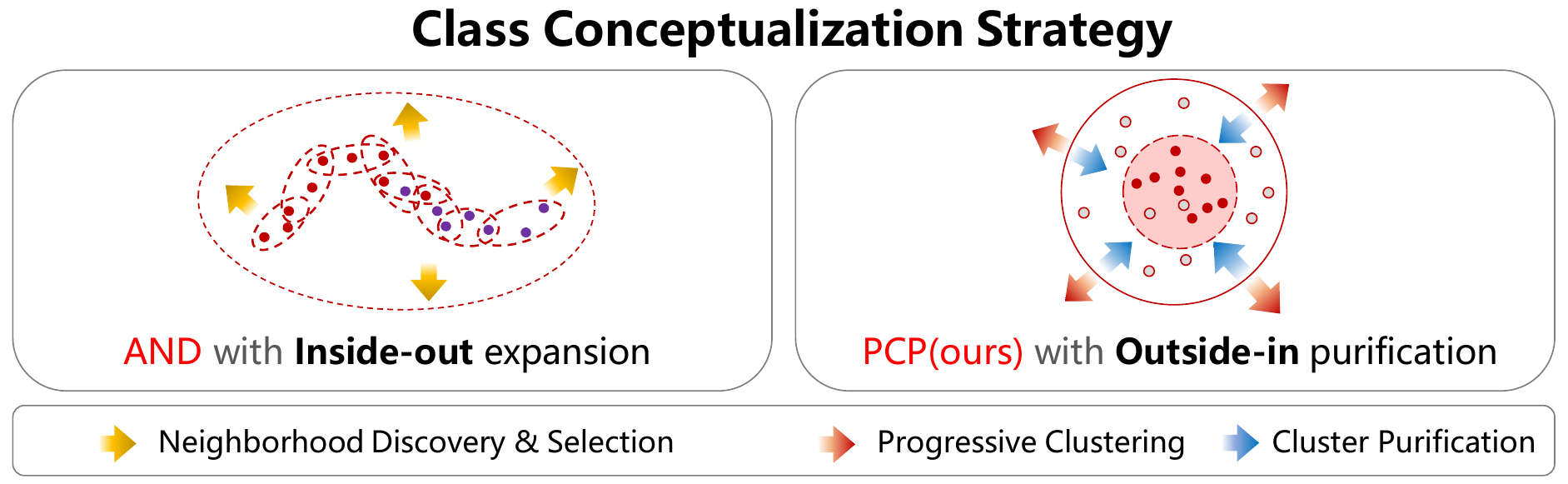}
     \caption[]{Comparison of class conceptualization strategies used by AND \cite{Huang_2019_ICML} and PCP.}
     \label{fig:discussion}
 \end{figure}
 
 % -----------------------
% \begin{figure*}[!t]
%      \centering
%      \includegraphics[width=12cm]{PCP_Fig4.png}
%      \caption{Performance comparison. (a) Comparison the effects of CP$_r$ and CP$_s$ under different filtering ratio. (b) Comparison of DC and PCP under different cluster numbers. (c) Comparison of AND and PCP for curriculum learning. $^*$ denotes with warm-up. %Detailed results show in Table.\ref{tab:table3}.
%      }
%      \label{fig:comparison}
%  \end{figure*}

\textbf{Reliability.} 
With PC and CP, PCP explores the underlying class information while mitigating the impact of clustering error. 
As shown in Fig.~\ref{fig:discussion}, PCP uses outside-in filtering strategy for class conceptualization, which generates reliable class consistent cluster by discarding noise samples. In contrast, AND \cite{Huang_2019_ICML} takes an inside-out expanding strategy to form class concept by merging neighborhoods. However, if the neighborhood size is large, it likely absorbs noise samples and drifts to be confused for feature learning; if the neighborhood size is small, the efficiency of model learning will be reduced, Table~\ref{tab:table4+}.

\begin{figure}[!t]
     \centering
     \includegraphics[width=1.0\columnwidth]{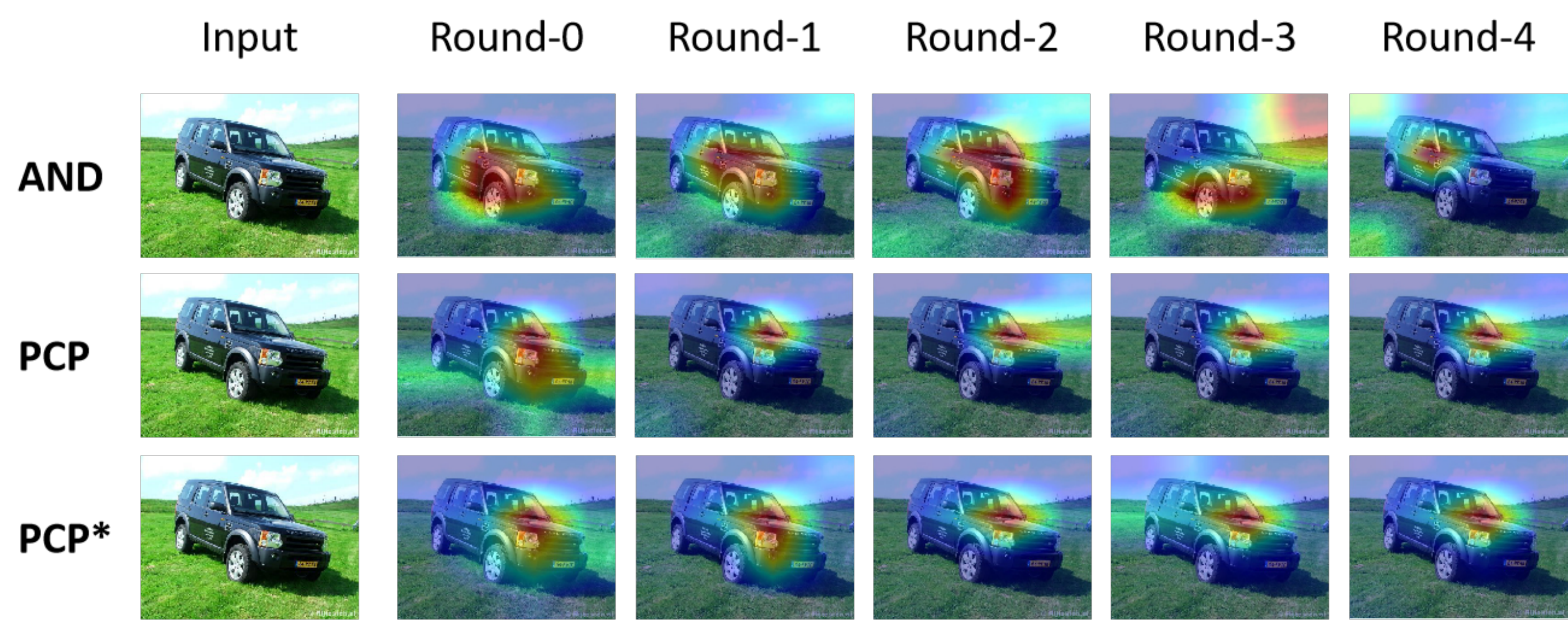}
     \caption{Visualization of features extracted by AND and PCP, which clearly shows that the features learned by PCP steadily focus on objects. $^*$ denotes with warm up.}
     \label{fig:vis_and_pcf}
 \end{figure}

\textbf{Stability.} 
With a simple-yet-effect clustering correction mechanism, PCP gets rid of dependence of approximating the number of classes of the training set from which other clustering based methods suffer a lot
%, Fig.~\ref{fig:comparison} (b).
%A voting function is utilized to take account historical clustering information, which improves the stability of the class boundary decision. 
%
It can be found that the learning procedure of PCP converges more stably, Fig.~\ref{fig:motivation} and Fig.~\ref{fig:vis_and_pcf}. 
%. until end of the training procedure.

{\bf Efficiency.} 
With the warm-up strategy, PCP extracts discriminative features in early training epochs, which not only improves the reliability and stability of feature learning but also results in fast convergence of training effect. Meanwhile, with quick decrease of cluster number, PCP is forced to efficiently perform class conceptualization. In contrast, AND sets the smallest neighbourhood size to avoid class drift, which limits its learning efficiency.
% Fig.~\ref{fig:comparison} (c). 

\section{Experiments}
%---------------------------------------
\setlength{\tabcolsep}{4pt}
\begin{table}
\begin{center}
\caption{Effects of the components in our approach with $k$NN classification accuracy. }
\label{tab:table1}
\begin{tabular}{ccccc}
\hline\noalign{\smallskip}
DC~\cite{Caron_2018_ECCV}  &PC &CP$_{r}$ &CP$_{s}$  &Acc  \\
\noalign{\smallskip}
\hline
\noalign{\smallskip}
$\surd$  &- &- &- &73.6 \\
      $\surd$  &$\surd$ &- &-  &76.9  \\
      $\surd$  &$\surd$ &$\surd$ &-  &78.9  \\
      $\surd$  &$\surd$ &$\surd$ &$\surd$ &81.6  \\
\hline
\end{tabular}
\end{center}
\end{table}
\setlength{\tabcolsep}{1.4pt}

%In this section, we first introduce the experimental settings. We then analyze the effects of the model components in PCP. Finally we evaluate the proposed PCP approach and compare it with other state-of-the-art methods.

\textbf{Datasets.} The learned feature representation is validated on image classification and object detection.  For image classification, we use CIFAR \cite{CIFAR} which contains 50,000/10,000 train/test images, and CIFAR10/CIFAR100 has 10/100 object classes with 6,000/600 images per class, all images with size 32 $\times$ 32. And IN-100 dataset which is selected from ImageNet~\cite{ImageNet-CVPR09} with 100 classes. For fine-grained classification, we use {CUB-200-2011} \cite{CUB} dataset which consists of 5994/ 5994 train/test images with 200 bird species. For object detection, we use {PASCAL VOC 2007} \cite{VOC} which contains 20 classes, 5011 images in trainval set and 4952 images in test set. 
%We evaluate object detection on the {PASCAL VOC 2007} dataset~\cite{VOC}, which consists of 5011 trainval images and 4952 test images over 20 object categories.

%----------------------------------------------
\setlength{\tabcolsep}{4pt}
\begin{table}
\begin{center}
\caption{Evaluation of CP$_{r}$ and CP$_{s}$ under different filtering ratio. }
\label{tab:table2}
\begin{tabular}{c|ccccccc}
\hline\noalign{\smallskip}
$\gamma$  &0 &0.3 &0.5 &0.7 &0.9 &0.95 &0.99 \\
\noalign{\smallskip}
\hline
\noalign{\smallskip}
CP$_{r}$ &76.9 &77.7 &78.9 &79.8 &81.4 &81.6 &81.7 \\
% CP$_{1-2}$ &80.3 &81.5 &81.6 &81.3 &81.1 &80.9 &80.6 \\
CP$_{r}$+CP$_{s}$ &80.3 &81.5 &81.6 &81.3 &81.1 &80.9 &80.6 \\
\hline
\end{tabular}
\end{center}
\end{table}
\setlength{\tabcolsep}{1.4pt}

%--------------------------------------------

\textbf{Experimental Setting.} Following the setting of \cite{Wu_2018_CVPR,Ye_2019_CVPR}, we initialize the learning rate as $0.03$ with the decreasing strategy that scaled-down by $0.1$ at $120$-th epoch and $0.01$ at $160$-th epoch. Then we set momentum to $0.9$, weight decay to $0.0005$, batch size to $128$, temperature parameter $\tau$ to $0.1$, and embedding feature space dimension to $128$ for network training. For fair comparison, we train PCP models with $200$ epochs (a round) as default, and choose $k = 200$ while implementing $k$NN evaluation. For the reliability of voting function for cluster purification, we implement cluster purification after $100$ epochs training and take $15$ epochs clustering results for voting function with two thresholds $0$ and $3.0$. For IS branch warm-up, the weight of loss in IS $t$ decrease from $0.8$ to $0.5$ during epoch $0$ to epoch $180$  continuously, and the weight of PCP with $1-t$. For multiple rounds training, we apply warm-up in the first round, the rest of rounds without warm-up and with the same decreasing strategy of learning rate. One-off AND is implemented with 2 rounds for the first round as warm-up. Unless stated otherwise, the reported data 
is implemented on CIFAR10 with ResNet18, and $k$NN performance adopted top-1 accuracy.
% DC is implemented under our loss function as show in Eq.~(\ref{lcluster}).

 %------------------------------------------------------
\begin{figure*}[t]
     \centering
     \includegraphics[width=1.6\columnwidth]{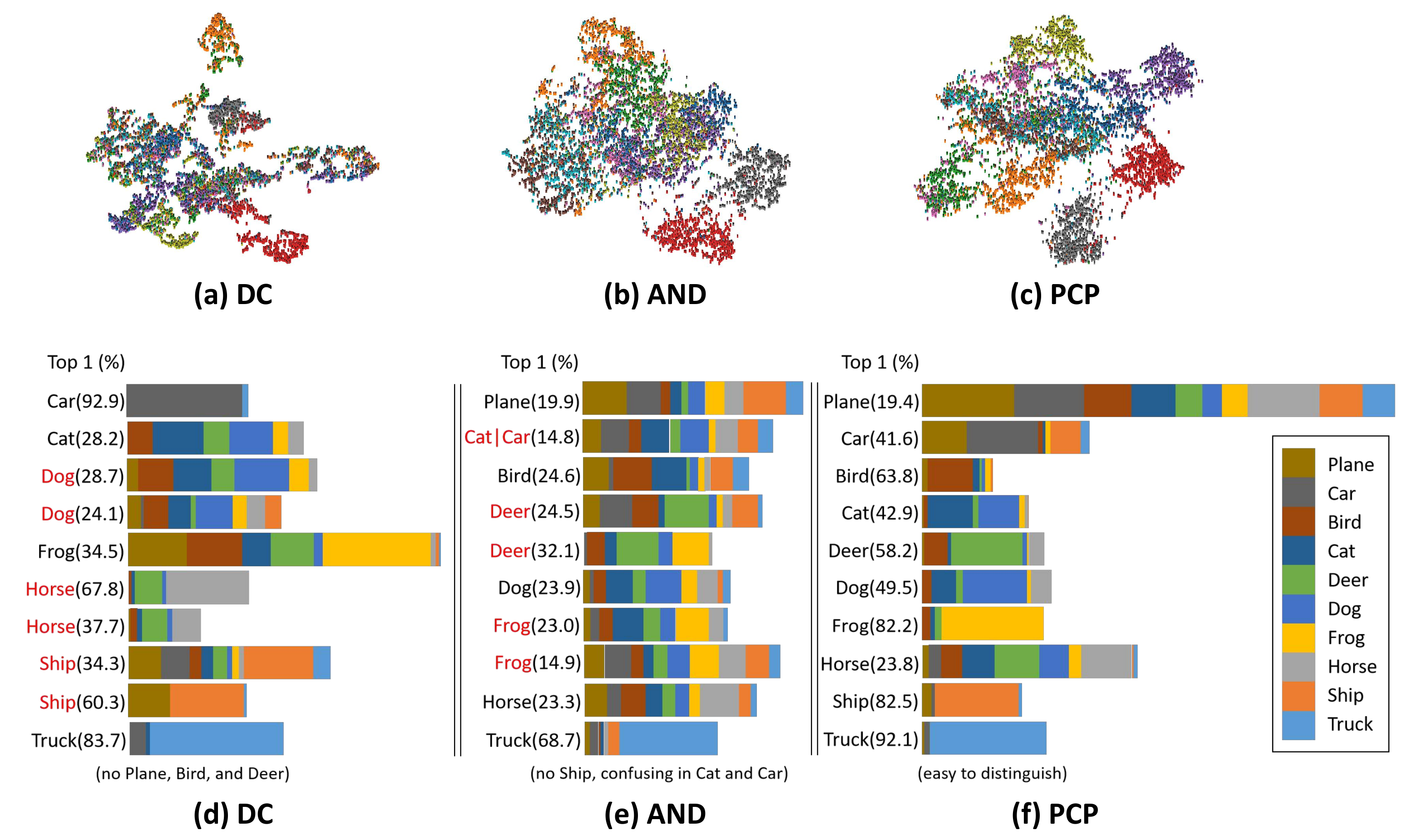}
     \caption{Visualization of 2-dimensional t-SNE distributions of the feature space (a-c) and its class statistics under $k$-means (k=10) clustering results (d-f) by DC , AND and PCP. (Best viewed in color)}
     \label{fig:cluster}
 \end{figure*}

\textbf{Evaluation Metrics.} We use linear classification and weighted $k$NN to evaluate features from different layers~\cite{Wu_2018_CVPR}. With linear classification, we implement classifier by designing a fully connected (FC) layer. We are aiming to optimize a FC layer by minimizing cross-entropy loss. With weighted $k$NN, for a test image $\hat{x}$, we calculate the cosine similarity $s_i$ = cos($v_i$, $\hat{v}$) for each of sample $x_i$ belonging to the train set, where $\hat{v}$ = $f_\theta$($\hat{x}$). The set of top-$k$ nearest neighbours denoted as $\mathcal{N}_k$ is then used to predict the class according to $\omega_c$ = $\sum_{i\in \mathcal{N}_k} \alpha_i \cdot 1 (c_i = c)$, where $\alpha_i = exp(s_i / \tau)$ and $\omega_c$ denotes the value of voting function that sample $\hat{x}$ belongs to class $c$. The result label of sample $\hat{x}$ is $c^*$ when $\omega_{c^*}$ = $\max$($\omega_0, \omega_1, \dots, \omega_{n-1}$), where $n$ denotes the number of categories in test set.

% \textbf{Compared Methods.} We compare PC with Split-Brain \cite{splitbrain_2017_CVPR}, Counting \cite{counting_2017_ICCV}, DC \cite{Caron_2018_ECCV}, IR \cite{Wu_2018_CVPR}, IS \cite{Ye_2019_CVPR} and AND \cite{Huang_2019_ICML} in the experiments.

%--------------------------------------

\subsection{Component Analysis and Discussion}
For fair comparison with DC, we drop cluster number (N$_{c}$) down to 100 to evaluate the effect of PCP components. 
% CP$_{r}$ denotes cluster purification with unreliable sample filtering, CP$_{s}$ denotes cluster purification with unstable sample filtering. 
In Table~\ref{tab:table1}, 
% Cluster as Label (CL) is similar to DC by taking cluster result as pseudo label, 
DC method achieves accuracy 73.6\% by taking cluster result as pseudo label,
we can see that PC improves the accuracy by 3.3\%, unreliable sample filtering of cluster purification (CP$_{r}$) gains 2\% and unstable sample filtering of cluster purification (CP$_{s}$) further gains 2.7\%.

%-----------------------------------------------

{\bf Cluster Purification.} 
In Table~\ref{tab:table2}, we explore the effect of CP on PCP
%with progressive clustering (PC)%
in detail.
% CP$_{r}$+CP$_{s}$ denotes a combination of cluster purification with unreliable and unstable sample filtering. 
As we can see that with the increase of the unreliable sample filtering ratio $\gamma$ from 0 to 0.99, the classification accuracy of PCP with only CP$_{r}$ consistently increases from 76.9\% to 81.7\%. This can be mainly attributed to that the supervision offered by clustering contains less noise. However, with higher $\gamma$, few samples remain in the cluster which limits the model to discovery inter-sample relationship. Further with CP$_{s}$, lower $\gamma$ leads to more noise samples, higher $\gamma$ leads to model close to IR, best performance appears in intermediate. PCP can not only keep the cluster size but also effectively correct the error assignments that rebounds the accuracy when $\gamma$ is low compared with only CP$_{r}$.
% Fig.~\ref{fig:comparison} (a). 
This further evaluates the effect of CP with no doubt. In the following experiments, we set $\gamma$ to 0.5 as default.

%-----------------------------------------
\setlength{\tabcolsep}{4pt}
\begin{table}
\begin{center}
\caption{Performance under different cluster numbers. $^*$ denotes with warm-up.}
\label{tab:table3}
\begin{tabular}{c|ccccccc}
\hline\noalign{\smallskip}
N$_{c}$  &10k &5k &3k &1k &100 &10 &5 \\
\noalign{\smallskip}
\hline
\noalign{\smallskip}
DC~\cite{Caron_2018_ECCV}   &82.9 &83.0 &82.1 &80.6 &73.6 &62.0 &56.9 \\
PCP  &81.7 &81.6 &81.8 &82.3 &81.6 &82.0 &81.8 \\
PCP$^{*}$ &84.1 &84.4 &84.3 &84.7 &83.9 &83.1 &83.2 \\
\hline
\end{tabular}
\end{center}
\end{table}
\setlength{\tabcolsep}{1.4pt}

{\bf Progressive Clustering.} 
As shown in Table~\ref{tab:table3}, the performance of DC drops dramatically from 82.9\% to 56.9\% with the cluster number (N$_{c}$) decreasing from 10,000 to 5 which is below the true class number of 10. In contrast, with the class conceptualization and the noise sample correction mechanism PC and CP, the performance of PCP is insensitive to N$_{c}$, which demonstrates the superior reliability and stability of PCP. With warm-up, PCP outperforms DC with a significant margin which is 4.1\% higher when N$_{c}$ is equal to 1k (set as default in the following experiments).
% , Fig.~\ref{fig:comparison} (b). 

%-----------------------------------------------------------------
\setlength{\tabcolsep}{4pt}
\begin{table}
\begin{center}
\caption{Comparison of AND and PCP (w or w/o warm-up) under different training rounds ($k$NN accuracy). $^*$ denotes with warm-up. N denotes neighborhood size.}
\label{tab:table4+}
\begin{tabular}{c|ccccc}
\hline
\noalign{\smallskip}
Round &0 &1 &2 &3 &4 \\
\noalign{\smallskip}
\hline
\hline
\noalign{\smallskip}
% DC &73.6 &77.3 &78.0 &78.4 &78.9 \\
% DC (100) &73.6 &77.3 &78.0 &78.4 &78.9 \\
% DC (1000) &80.6 &84.3 &86.0 &86.0 &86.0 \\
% \hline
   AND~\cite{Huang_2019_ICML} (N = 1)  &80.0 &83.9 &85.0 &85.9 &86.3  \\
    % AND~\cite{Huang_2019_ICML} (N = 5)  &80.6 &84.2 &85.3 &85.6 &85.9  \\
      AND~\cite{Huang_2019_ICML} (N = 5)  &80.0 &83.8 &85.0 &85.6 &85.8  \\
    AND~\cite{Huang_2019_ICML} (N = 10)  &80.0 &83.8 &84.9 &85.1 &85.1 \\
   AND~\cite{Huang_2019_ICML} (N = 20)  &80.0 &83.3 &84.3 &84.6 &84.6 \\
\hline
     PCP (Ours)   &82.3 &84.8 &85.3 &86.0 &86.0 \\
     PCP$^*$ (Ours) &{\bf 84.7} &{\bf 86.4} &{\bf 86.7} &{\bf 87.0} &{\bf 87.3} \\
\hline
\end{tabular}
\end{center}
\end{table}
\setlength{\tabcolsep}{1.4pt}

%---------------------------------------------------------------------
\setlength{\tabcolsep}{4pt}
\begin{table}
\begin{center}
\caption{Comparison of classification accuracy ($k$NN) on CIFAR10. F-S denotes fully supervised. $^*$ denotes the performance produced by our implementation. $^+$ denotes 5 rounds training. The compared results of IR/IS/AND are directly transcribed from their references.
}
\label{tab:c_res18}
\begin{tabular}{c|ccccc}
\hline\noalign{\smallskip}
 Model  &Random &F-S$^*$ &DC$^*$~\cite{Caron_2018_ECCV} &IR~\cite{Wu_2018_CVPR} &IS~\cite{Ye_2019_CVPR}  \\
\noalign{\smallskip}
\hline
\noalign{\smallskip}
Acc &32.1 &93.1 &80.6 &80.8 &83.6 \\
\hline
\hline\noalign{\smallskip}
Model  &Random  &AND~\cite{Huang_2019_ICML} &PCP &AND$^+$ &{\bf PCP$^+$} \\
\noalign{\smallskip}
\hline
\noalign{\smallskip}
Acc &32.1 &84.2 &84.7 &86.3 &{\bf 87.3} \\
\hline
\end{tabular}
\end{center}
\end{table}
\setlength{\tabcolsep}{1.4pt}
 
 {\bf Curriculum Learning.} 
To enhance discriminative power of learned features, AND derives curriculum learning with five training rounds to select class-consistent neighbourhoods. In Table~\ref{tab:table4+} we can see that the convergence of PCP is faster and the performance is higher than those of AND, and PCP with warm-up further converges faster and performs better which show the superior efficiency of PCP.
% Fig.~\ref{fig:comparison} (c). 
% DC shows comparable accuracy with AND and PCP, however, DC is sensible for cluster number with a margin about 7\%. 
From the visualization in Fig.~\ref{fig:vis_and_pcf}, that the features learned by PCP steadily focus on objects shows the superior stability of PCP.
% PCP with warm-up improves the performance by the margin of 1\% $\thicksim$ 2\%, and the extracted feature is also similar to PCP.

%--------------------------------------------------------------------------------------
\textbf{Class Conceptualization.} 
To show the effect of class conceptualization, we visualize the feature distribution of the whole test set of CIFAR10 by embedding the 128-dimensional feature space into a 2-dimensional space by t-SNE~\cite{tsne}. In Fig.~\ref{fig:cluster} (a-c), both DC and PCP models are trained with N$_{c}$ = 10 and we can see that the feature representations learned by DC and AND are less discriminative as many features with different classes mixed up. In contrast, the features extracted by PCP are better assigned to different clusters. 

 %------------------------
\setlength{\tabcolsep}{4pt}
\begin{table}
\begin{center}
\caption{Evaluation on CIFAR10 and CIFAR100 with AlexNet by performing linear classifier on the features from $conv5$, and $k$NN from $FC$. F-S denotes fully supervised. $^*$ denotes our rerunning. AND in our implementation has two rounds (one-off) while PCP with one round.}
\label{tab:alex_compare}
\begin{tabular}{c|cc|cc}
\hline\noalign{\smallskip}
 Classifier &\multicolumn{2}{c|}{Weighted $k$NN ($FC$)} &\multicolumn{2}{c}{Linear Classifier ($conv5$)}\\
 \hline
\noalign{\smallskip}
Dataset   &CIFAR10 &CIFAR100 &CIFAR10 &CIFAR100  \\

\hline
\hline
\noalign{\smallskip}
% Split-Brain~\cite{splitbrain_2017_CVPR} &11.7 &1.3 &67.1 &39.0 \\
    %   Counting~\cite{counting_2017_ICCV} &41.7 &15.9 &50.9 &18.2 \\
    %   DC &62.3 &22.7 &77.9 &41.9 \\
    %   IR &60.3 &32.7 &70.1 &39.4 \\
      
    %   AND &74.8 &41.5 &77.6 &47.9  \\
     % \hline
      DC$^*$~\cite{Caron_2018_ECCV} &70.3 &27.4 &77.1 &44.0 \\
      IR$^*$~\cite{Wu_2018_CVPR} &68.1 &39.6 &76.6 &49.5 \\
      IS$^*$~\cite{Ye_2019_CVPR} &76.4 &46.3 &78.7 &51.2 \\
      AND$^*$~\cite{Huang_2019_ICML} &76.1 &44.2 &79.2 &52.8 \\
      
      {\bf PCP (Ours)} &{\bf 77.1} &{\bf 48.4} &{\bf 79.9} &{\bf 53.0} \\
      \hline
       F-S &91.9 &69.7 &91.8 &71.0\\
\hline
\end{tabular}
\end{center}
\end{table}
\setlength{\tabcolsep}{1.4pt}

To quantitatively show the superiority of PCP, we cluster learned features to $10$ clusters, then identify the class pseudo-label of each cluster by the category which the majority of features in the cluster belong to. As shown in Fig.~\ref{fig:cluster} (d-f), PCP forms 10 clusters dominated by samples from 10 different classes. For DC, 3 categories ($Plane$, $Bird$ and $Deer$) are missed in that each of the 3 categories ($Dog$, $Horse$ and $Ship$) dominates two clusters
respectively. AND misses the class of $Ship$, and shows same proportions in  $Cat$ and $Car$. We can easily distinguish each category in PCP
%of Fig.~\ref{fig:cluster} (d - f)%
while DC and AND look slightly confusing. So far we can argue that better class conceptualization is of essential importance to improve the performance of unsupervised clustering methods.

%---------------------------------------------------------------------
% \setlength{\tabcolsep}{4pt}
% \begin{table}
% \begin{center}
% \caption{Evaluation on IN-10/IN-50/IN-100 datasets with AlexNet by performing linear classifier(LC) on the features from $conv5$, and $k$NN from $FC$. }
% \label{tab:pcp-moco}
% \begin{tabular}{c|cc|cc}
% \hline\noalign{\smallskip}
% Model &\multicolumn{2}{c|}{MoCo~\cite{moco2020}} &\multicolumn{2}{c}{{\bf PCP(Ours)}}\\
%  \hline
% \noalign{\smallskip}
% Classifier   &$k$NN &LC &$k$NN &LC  \\

% \hline
% \hline
% \noalign{\smallskip}
% % Split-Brain~\cite{splitbrain_2017_CVPR} &11.7 &1.3 &67.1 &39.0 \\
%     %   Counting~\cite{counting_2017_ICCV} &41.7 &15.9 &50.9 &18.2 \\
%     %   DC &62.3 &22.7 &77.9 &41.9 \\
%     %   IR &60.3 &32.7 &70.1 &39.4 \\
      
%     %   AND &74.8 &41.5 &77.6 &47.9  \\
%      % \hline
%       IN-10 &46.0 &67.4 &77.6 &80.6 \\
%     %   IN-10 &49.0 &64.6 &77.6 &80.6 \\
%       IN-50 &59.0 &63.4 &62.3 &65.6 \\
%       IN-100 &50.1 &55.4 &50.7 &56.9 \\
      
% \hline
% \end{tabular}
% \end{center}
% \end{table}
% \setlength{\tabcolsep}{1.4pt}

%---------------------------------------------------------------------
\setlength{\tabcolsep}{4pt}
\begin{table}
\begin{center}
\caption{Comparison with MoCo~\cite{moco2020} on ImageNet subset IN-100 with AlexNet by performing linear classifier(LC) on the features from $conv5$, and $k$NN from $FC$. }
\label{tab:pcp-moco}
\begin{tabular}{c|cc}
\hline\noalign{\smallskip}

\noalign{\smallskip}
Classifier   &$k$NN ($FC$) &LC ($conv5$)  \\

\hline
\hline
\noalign{\smallskip}
% Split-Brain~\cite{splitbrain_2017_CVPR} &11.7 &1.3 &67.1 &39.0 \\
    %   Counting~\cite{counting_2017_ICCV} &41.7 &15.9 &50.9 &18.2 \\
    %   DC &62.3 &22.7 &77.9 &41.9 \\
    %   IR &60.3 &32.7 &70.1 &39.4 \\
      
    %   AND &74.8 &41.5 &77.6 &47.9  \\
     % \hline
    %   MoCo~\cite{moco2020} &46.0 &67.4 &59.0 &63.4 &50.1 &55.4 \\
    % %   IN-10 &49.0 &64.6 &77.6 &80.6 \\
    %   {\bf PCP(Ours)} &77.6 &80.6 &62.3 &65.6 &50.7 &56.9 \\
    MoCo~\cite{moco2020} &50.1 &55.4 \\
    %   IN-10 &49.0 &64.6 &77.6 &80.6 \\
      {\bf PCP(Ours)}  &50.7 &56.9 \\
      
\hline
\end{tabular}
\end{center}
\end{table}
\setlength{\tabcolsep}{1.4pt}
% ------------------------
% \setlength{\tabcolsep}{4pt}
% \begin{table}
% \begin{center}
% \caption{Evaluation on ImageNet100 with AlexNet by performing linear classifier(LC) on the features from $conv5$, and $k$NN from $FC$. The queue size of MoCo is set to 6528.}
% \label{tab:pcp-moco}
% \begin{tabular}{ccc}
% \hline\noalign{\smallskip}
% Classifier &$k$NN($FC$)  &LC($conv5$)\\
% \noalign{\smallskip}
% \hline
% \noalign{\smallskip}
% MoCo~\cite{moco2020}  &43.5 &49.5\\
% PCP  &50.7 &56.9\\
% %&50.1 
% \hline
% \end{tabular}
% \end{center}
% \end{table}
% \setlength{\tabcolsep}{1.4pt}
 %--------------------------------------------------------------------

\setlength{\tabcolsep}{4pt}
\begin{table}
\begin{center}
\caption{Comparison of fine-grained classification performance. $^*$ denotes our rerunning. PCP is implemented with one round.}
\label{tab:cub-fine-grained}
\begin{tabular}{c|cccccc}
\hline\noalign{\smallskip}
Model  &Random$^*$ &IR~\cite{Wu_2018_CVPR} &DC$^*$~\cite{Caron_2018_ECCV} &AND~\cite{Huang_2019_ICML} &{\bf PCP (Ours)}\\  %&F-S$^*$
\noalign{\smallskip}
\hline
\noalign{\smallskip}
Acc  &2.6 &11.6 &13.1 &14.1 &{\bf 16.9}\\
%&50.1 
\hline
\end{tabular}
\end{center}
\end{table}
\setlength{\tabcolsep}{1.4pt}

\subsection{Image Classification}

To further evaluate the proposed PCP approach, we conduct experiments on image classification and fine-grained classification to compare with baseline methods.
The backbone is ResNet18 and the classification is evaluated by $k$NN. The results of IR/IS/AND are directly transcribed from the corresponding references.
DC is reimplemented with 1000 clusters. As shown in Table~\ref{tab:c_res18}, PCP model achieves the best performance of 84.7\% accuracy, and after 5 rounds the accuracy of 87.3\% outperforms AND with 1\%. Considering the fully-supervised accuracy is only 93.1\%, the performance of PCP is extraordinary under unsupervised setting.

%As show in Table~\ref{tab:c_res18}, with uniformly implemented with a ResNet18 backbone and by $k$NN evaluation, DC is reimplemented under our loss function by clustering 1000 clusters. PCP model achieves the best performance of 84.7\% accuracy and 5 rounds of 87.3\% accuracy compared with stat-of-the-art methods on CIFAR10 dataset.
%

%
For generalization verification, we then compare image classification accuracy on CIFAR10 \& CIFAR100 with a standard AlexNet backbone attached by a weighted $k$NN ($FC$ layer) and a linear classifier ($conv5$ layer). 
The figures in Table~\ref{tab:alex_compare} clearly show that PCP outperforms other methods, and especially with a significant margin for CIFAR100 evaluated by $k$NN.

We further compare our method with MoCo~\cite{moco2020} (rerun the code \url{https://github.com/facebookresearch/moco}, 
default batch size as $128$, queue size as $6528$, and $2$ GeForce RTX 2080Ti GPUs) on ImageNet100 subset IN-100 under the same settings, the performance 
of PCP is higher than MoCo with 0.6\% for $k$NN top-1 accuracy and 1.5\% for linear classifier accuracy, Table~\ref{tab:pcp-moco}.

%---------------------------------------------------------------------

{\bf Fine-grained Classification.} 
PCP is further evaluated on the more challenging fine-grained classification task. We follow the settings with N$_{c}$ =  $500$ and batch size as $64$ on the CUB-200-2011 dataset. As shown in Table~\ref{tab:cub-fine-grained}, PCP significantly outperforms IR by 5.3\%, DC by 3.8\% and AND by 2.5\%.

%----------------------------------------
\setlength{\tabcolsep}{4pt}
\begin{table}
\begin{center}
\caption{Comparison of object detection performance. AND in our implementation has two rounds (one-off) while PCP with one round.}
\label{tab:detection2}
\begin{tabular}{c|cccccc}
\hline\noalign{\smallskip}
Model  &Random &DC~\cite{Caron_2018_ECCV} &IR~\cite{Wu_2018_CVPR}  &AND~\cite{Huang_2019_ICML} &{\bf PCP (Ours)} &F-S\\
\noalign{\smallskip}
\hline
\noalign{\smallskip}
mAP &0.5 &27.8 &30.6 &36.9 &{\bf 39.8} &46.1 \\
\hline
\end{tabular}
\end{center}
\end{table}
\setlength{\tabcolsep}{1.4pt}

\subsection{Object Detection}
Object detection experiments are conducted to evaluate the effectiveness of the features learned by PCP for down-stream tasks.
We implement Faster R-CNN \cite{fasterrcnn} with a ResNet18 backbone pre-trained on the CIFAR100 dataset and fine-tuned on the PASCAL VOC 2007 dataset. 

For pretrained ResNet18 backbone, the first residual stage and all the batch normalization layers are frozen.
Among the overall fine-tuning 30 epochs, the learning rate is first set to 0.001 and decreased 10 times after 25 epochs, and the image batch size is fixed as 8. 
Compared with the randomly initialized feature model, which is hard to converge, SOTA models significantly improve the $m$AP performance, Table~\ref{tab:detection2}. 
PCP achieves 39.8\% mAP which outperforms AND by 2.9\%. 
This shows the effectiveness of the features learned by PCP for down-stream tasks.

%-------------------------------------------------------------------------------------------------
\section{Conclusion}

In this paper, we proposed a simple-yet-effective Progressive Cluster Purification (PCP) method for unsupervised feature learning. To alleviate the impact of noise samples, we elaborately designed the Progressive Clustering (PC) strategy to gradually expand the cluster size consistently with the growth of the model representation capability and the Cluster Purification (CP) mechanism to reduce unreliable and unstable noise samples in each cluster to a significant extent. Resultantly, PCP mitigates the dependency of the prior knowledge of the cluster number and is able to reliably, stably and efficiently learn discriminative and representative features.  
With warm-up training strategy, PCP avoids network focusing on low-level feature for early clustering. Extensive experiments on classification and detection benchmarks demonstrated that the proposed PCP approach has improved the classical clustering method and provided a fresh insight into the unsupervised learning problem.

% conference papers do not normally have an appendix

% use section* for acknowledgment
%\section*{Acknowledgment}

%The authors would like to thank...

% trigger a \newpage just before the given reference
% number - used to balance the columns on the last page
% adjust value as needed - may need to be readjusted if
% the document is modified later
%\IEEEtriggeratref{8}
% The "triggered" command can be changed if desired:
%\IEEEtriggercmd{\enlargethispage{-5in}}

% references section

% can use a bibliography generated by BibTeX as a .bbl file
% BibTeX documentation can be easily obtained at:
% http://mirror.ctan.org/biblio/bibtex/contrib/doc/
% The IEEEtran BibTeX style support page is at:
% http://www.michaelshell.org/tex/ieeetran/bibtex/
%\bibliographystyle{IEEEtran}
% argument is your BibTeX string definitions and bibliography database(s)
%\bibliography{IEEEabrv,../bib/paper}
%
% <OR> manually copy in the resultant .bbl file
% set second argument of \begin to the number of references
% (used to reserve space for the reference number labels box)

\newpage
% \input{appendix}

% that's all folks
\end{document}